\documentclass[letterpaper]{article}
\usepackage{aaai19}
\usepackage{times}
\usepackage{helvet}
\usepackage{courier}
\usepackage[hyphens]{url} 
\usepackage{graphicx}
\usepackage{graphicx}
\usepackage{xparse}
\usepackage{amsmath}
\usepackage{amsthm}
\usepackage{amssymb}
\usepackage{xspace}
\usepackage{relsize}
\usepackage{multirow}
\usepackage{comment}

\newcommand{\subsubparagraph}[1]{}

\makeatletter
\let\@myref\ref

\newcommand{\refsec}[1]{Sec.\,\@myref{#1}}
\newcommand{\refseq}[1]{Sec.\,\@myref{#1}}
\newcommand{\refig}[1]{Fig.\,\@myref{#1}}
\newcommand{\refigs}[2]{Fig.\,\@myref{#1}-\@myref{#2}}

\newcommand{\reftbl}[1]{Table \@myref{#1}}
\newcommand{\refstep}[1]{Step \@myref{#1}}
\newcommand{\refalgo}[1]{Algorithm \@myref{#1}}
\newcommand{\refchap}[1]{Chapter \@myref{#1}}
\newcommand{\reflst}[1]{List \@myref{#1}}
\newcommand{\refeq}[1]{Eq. \@myref{#1}}

\newcounter{list}[section]

\makeatother

\newcommand{\braces}[1]{{\left\{#1\right\}}}
\newcommand{\parens}[1]{{\left(#1\right)}}

\newcommand{\lsota}{state-of-the-art\xspace}  

\newcommand{\defun}[1]{%
\makeatletter
\expandafter\def\csname the#1\endcsname{\text{\it #1}}
\expandafter\def\csname #1\endcsname ##1{\csname the#1\endcsname\left(##1\right)}%
\makeatother
}

\newcommand{\defsetop}[2]{%
\makeatletter
 \expandafter\def\csname #1\endcsname ##1##2##3{%
  \expandafter\def\csname #1arg\endcsname{##1}%
  \expandafter\def\csname #1set\endcsname{##2}%
  \expandafter\def\csname #1cond\endcsname{##3}%
  \braces{##1##2\mid #2 ##3}%
 }%
\makeatother%
}

\defun{precond}
\defun{condition}
\defun{params}
\defun{type}
\defun{proc}
\defun{cost}

\defun{push}
\defun{pref}

\defun{init}
\defun{goal}
\defun{objects}
\defun{task}

\defun{parent}
\defun{plateau}

\defsetop{filter}{}
\defsetop{map}{}

\makeatletter
\NewDocumentCommand{\todo}{s O{} m}{%
  \IfBooleanTF#1%
    {\@todos{#2}{#3}}%
    {\@todons{#2}{#3}}}
\newcommand{\@todons}[2]{}
\newcommand{\@todos}[2]{}
\makeatother

\def\_{\\[-0.3em]}

\makeatletter

\newcommand{\newheuristic}[2]{%
 \def#1{%
  \ifmmode%
  h^\text{#2}\xspace%
  \else%
  \text{#2}\xspace%
  \fi%
 }%
}

\newheuristic{\lmcut}{LMcut}
\newheuristic{\mands}{M\&S}
\newheuristic{\pdb}{PDB}
\newheuristic{\ff}{FF}
\newheuristic{\ce}{CEA}
\newheuristic{\cg}{CG}
\newheuristic{\ad}{add}
\newheuristic{\lc}{LC}

\newcommand{\newUnitCostHeuristic}[2]{%
 \def#1{%
  \ifmmode%
  \hat{h}^\text{#2}\xspace%
  \else%
  \text{#2}\xspace%
  \fi%
 }%
}

\newUnitCostHeuristic{\lmcuto}{LMcut}
\newUnitCostHeuristic{\mandso}{M\&S}
\newUnitCostHeuristic{\ffo}{FF}
\newUnitCostHeuristic{\ceo}{CEA}
\newUnitCostHeuristic{\cgo}{CG}
\newUnitCostHeuristic{\ado}{add}
\newUnitCostHeuristic{\gco}{GoalCount}
\newUnitCostHeuristic{\lco}{LC}

\makeatother

\def\latentplanner{Latplan\xspace}

\newcommand{\before}{pre}
\newcommand{\after}{suc}

\usepackage{amsmath,amsfonts,bm}

\def\eqref#1{equation~\ref{#1}}

\def\1{\bm{1}}

\def\va{{\bm{a}}}

\def\vg{{\bm{g}}}

\def\vx{{\bm{x}}}

\def\vz{{\bm{z}}}

\DeclareMathAlphabet{\mathsfit}{\encodingdefault}{\sfdefault}{m}{sl}
\SetMathAlphabet{\mathsfit}{bold}{\encodingdefault}{\sfdefault}{bx}{n}

\newcommand{\R}{\mathbb{R}}

\author{Masataro Asai \\ IBM Research \\ MIT-IBM Watson AI Lab}

\title{Unsupervised Grounding of \\ Plannable First-Order Logic Representation from Images}

\begin{document}
\maketitle
\begin{abstract}
Recently, there is an increasing interest in obtaining the relational structures
of the environment in the Reinforcement Learning community.
However, the resulting ``relations'' are not
the discrete, logical predicates compatible with the symbolic reasoning such as classical planning or goal recognition.
Meanwhile, \latentplanner \cite{Asai2018} 
bridged the gap between deep-learning perceptual systems and symbolic classical planners.
One key component of the system is a Neural Network called State AutoEncoder (SAE),
which encodes an image-based input into a propositional representation compatible with classical planning.
To get the best of both worlds,
we propose First-Order State AutoEncoder,
an unsupervised architecture for grounding the first-order logic predicates and facts.
Each predicate models a relationship between objects by
taking the interpretable arguments and returning a propositional value.
In the experiment using 8-Puzzle and a photo-realistic Blocksworld environment,
we show that
(1) the resulting predicates capture the interpretable relations (e.g., spatial),
(2) they help to obtain the compact, abstract model of the environment,
and finally,
(3) the resulting model is compatible with symbolic classical planning.
\end{abstract}

\section{Introduction}

Recent success in the latent space classical planning \cite[Latplan]{Asai2018} shows a promising direction for
connecting the neural perceptual systems and the symbolic AI systems.
Latplan is a straightforward system built upon a \lsota Neural Network (NN) framework (Keras, Tensorflow) and Fast Downward classical planner \cite{Helmert04}. 
It builds a set of propositional state representation from the raw observations (e.g., images) of the environment,
which can be used for classical planning as well as goal recognition \cite{amado2018goal}.
However,
Latplan still contains many rooms for improvements
in terms of the interpretability and the scalability which are trivially available in the symbolic systems.
An instance of such limitations of Latplan is that the reasoning is performed on a propositional level,
missing the ontological commitment of the first-order logic (FOL) that \emph{the world comprises objects and their relations} \cite{russell1995artificial}.

\begin{figure}[tb]
 \centering
 \includegraphics[width=\linewidth]{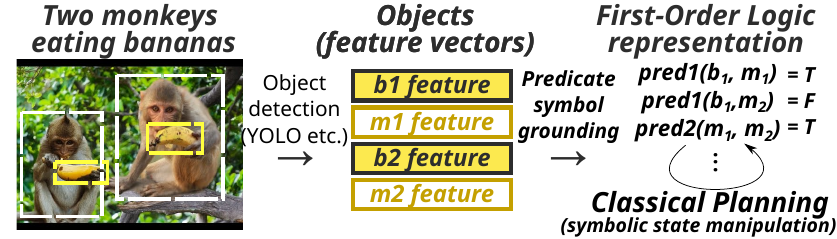}
 \caption{
 \emph{Predicate symbol grounding} (PSG) process for identifying the
 predicates and obtaining the first-order logic (FOL) representation of
 the environment for symbolic reasoning.  In this example, an anonymous
 binary predicate \textit{pred$_1$} can be interpreted by humans as
 something like \emph{eating}$(\textit{object},\textit{subject})$.
 }
 \label{predicate-symbol-grounding}
\end{figure}

FOL is a \emph{structured} representation, which offers some extent of interpretability
compared to the \emph{factored} representation of the propositional logic formula \cite{russell1995artificial}.
Even if the predicate symbols discovered by a \emph{Predicate Symbol Grounding} system (\refig{predicate-symbol-grounding})
are machine-generated anonymous symbols (not the human-originated symbols assigned by manual tagging),
the structures help humans interpret the meaning of the relations from the several instances of
the argument list (objects) that make the predicate true.
For example, when two propositions \textit{pred}(0,1) and \textit{pred}(1,2) are true,
we can guess the meaning of \textit{pred} as \texttt{+1},
or given \textit{pred}(\textit{monkey}, \textit{banana}) being true,
the meaning of \textit{pred} would be something like \emph{eating} or \emph{holding}.
This is impossible in a propositional representation where only the variable indices and the truth values are known.

In this paper, we propose First-Order State AutoEncoder (FOSAE, \refig{fig:fosae}), a NN architecture which, given the feature vectors of the objects in the environment,
automatically learns to identify a set of predicates (relations)
as well as to select the appropriate objects as the arguments for the predicates.
The resulting representation is compatible with classical planning.
We do not address the object recognition problem, whose task is to extract the object entities from a raw observation.
We instead assume that an external system has already extracted the objects and converted them into the feature vectors,
given the recent success of object detection methods like YOLO \cite{redmon2016you} in image processing.
While FOSAE is in principle data-format (e.g., images, text) independent, we focus on the visual input in this paper.

FOSAE provides a higher-level generalization and the more compact model by
adding a constraint that the extracted relations are common to multiple tuples of objects.
Ideally, predicates model the commonalities between the multiple instantiations
of its arguments, rather than rote learning some unrelated combinations.
In order to discover such predicates,
our framework ensures that a single predicate is applied to the different arguments \emph{within} the same observation.
Otherwise, the network may choose to apply them to the same or the very narrow combinations of arguments in every observation,
resulting in an inflexible predicate that merely remembers some combinations.
Since the model utilizes the same weights multiple times over the different arguments,
this also reduces the number of weight parameters required to model the environment.

The rest of this paper is organized as follows. 
\refsec{related} reviews and discusses the issues in the existing work on the relational structures for NNs.
\refsec{background} explains Latplan, the key target of the enhancement
proposed in this paper, as well as the backgrounds for NNs such as 
\emph{attention} that plays a key role in extracting the \emph{argument
list} of the predicates.
\refsec{requirements} discusses the process
for learning the first-order logic representation of the environment,
and \refsec{fosae} proposes First-Order State AutoEncoder (FOSAE), 
the architecture that performs the process.
In \refsec{experiments-8puzzle}, we conduct experiments on a toy 8-puzzle domain to address
the interpretability and generalization of FOSAE
and its compatibility to symbolic PDDL planning systems.
In \refsec{experiments-blocks}, we demonstrate the ability of FOSAE 
on a photo-realistic Blocksworld domain.

\section{Background}
\label{related}

One of the early work on learning the relations between object symbols
includes Linear Relational Embedding \cite{paccanaro2001learning},
which is learned from the tuples of relations and objects in a supervised manner.
Supervision makes the resulting representation parasitic to human knowledge; thus it does not fully solve the knowledge acquisition bottleneck.
Also, their approach is limited to binary relations.

Recently, there are increasing interest in the effectiveness of finding ``relations'' in Deep Reinforcement Learning \cite[DRL]{dqn} community.
Deep Symbolic RL \cite{garnelo2016towards} showed that a hand-crafted ``common sense prior'' (e.g., proximity) accelerates DRL.
Relation Networks \cite{santoro2017simple} combine two elements in the feature maps of a convolutional neural network \cite{lecun2015deep} and output real values.
Deep Relational RL \cite{zambaldi2018relational} combines Relation Networks and attention-based message passing.
Relational Inductive Bias \cite{battaglia2018relational} shows that DRL is enhanced by a hand-crafted, explicit graph representation of the input and the Graph Neural Networks \cite{scarselli2009graph}.
Relational Neural Expectation Maximization \cite{van2018relational} enumerates pairs of objects
to model the relations in the environment for common-sense physics modeling.
Neural theorem proving systems \cite{rocktaschel2017end,sourek2018lifted,manhaeve2018deepproblog} learn to reason about logical relations.
In this paper, we address the following issues in these work:

\textbf{Human Supervision.}
Providing a relational dataset as an input (as in \cite{battaglia2018relational} and neural theorem proving),
or a probabilistic logic program containing predicate symbols which defines a network, 
exhibits the knowledge acquisition bottleneck
as the predicates are grounded by humans and thus the system relies on human knowledge.

\textbf{Compatibility with the symbolic systems.}
Relational structures in existing work do not return explicit boolean values
even when the environment is deterministic, fully observable and discrete.
This makes them incompatible with symbolic systems such as classical planners or goal recognition.
Ideally, systems should guarantee that a discrete environment is represented in a discrete form,
and numeric variables (such as those handled by numeric planner) should be introduced only when necessary.

\textbf{Interpretability.}
Some networks use real-valued soft attentions (probability) to model the objects that take part in a relation,
which are similar to the predicate arguments.
However, the relations resulted from soft attentions are hard to interpret
due to the ambiguity, e.g., ``Bob has-a `50\% dog and 50\% cat''' in a ``has-a'' relation.
Continuous outputs of the relational structures are also difficult to interpret.

\textbf{Scalability for higher arities.}
Some work assumes the binary relations and enumerates $O(N^2)$ pairs for $N$ objects.
The explicit structure is impractical for larger arities $A$ because the network size $O(N^A)$ increases exponentially.

Furthermore, \textbf{empirical, direct evidence} that the ``relations'' are
indeed necessary for modeling the environment is missing from the literature.
In some previous work,
relational structures are embedded inside a reinforcement learning framework,
and they showed that such structures had improved the RL performance.
However, this is merely a necessary condition rather than a sufficient condition because,
in model-free RL, the policy and the representation are learned at once,
making it hard to tell whether
(1) the network is just numerically faster to converge (similar to skip connections or other techniques) or
(2) it models the environment better and \emph{therefore} it learns the good policy.

\section{Preliminaries}
\label{background}

We denote an array (either a vector or a matrix) in bold, such as $\vx$, and denote its elements or rows
with a subscript, e.g., when $\vx\in \R^{n\times m}$, the second row is $\vx_2 \in \R^m$.

\textbf{Symbol grounding} is an unsupervised process of establishing a mapping
from huge, noisy, continuous, unstructured inputs
to a set of compact, 
discrete, identifiable (structured) entities, i.e., symbols \cite{harnad1990symbol,taddeo2005solving,Steels2008,Asai2018}.
PDDL \cite{McDermott00} has six kinds of symbols: Objects, predicates, propositions, actions, problems, and domains.
Each type of symbol requires its own mechanism for grounding.
For example, the large body of work in the image processing community on recognizing 
objects (e.g., faces) and their attributes (male, female) in images, or scenes in videos (e.g., cooking)
can be viewed as corresponding to grounding the object, predicate and action symbols, respectively.

\textbf{Classical planners} such as FF \cite{hoffmann01} or
FastDownward \cite{Helmert04} takes a PDDL model as an input, which
specifies the state representation, the initial state, the goal
condition and the transition rules in the form of the first-order logic
formula, and returns an action sequence that reaches the goal state from the initial state.
In contrast,
\textbf{\latentplanner} \cite{Asai2018} is a framework for
\emph{domain-independent image-based classical planning}.
It learns the state representation as well as the transition rules
entirely from the image-based observation of the environment with deep neural networks
and solves the problem using a classical planner.
The system was shown to solve various puzzle domains, such as 8-puzzles or Tower of Hanoi,
which are presented in the form of noisy, continuous visual depiction of the environment.
\latentplanner addresses two of the six types of symbols, 
namely propositional and action symbols.

\begin{figure}[tb]
 \centering
 \includegraphics[width=\linewidth]{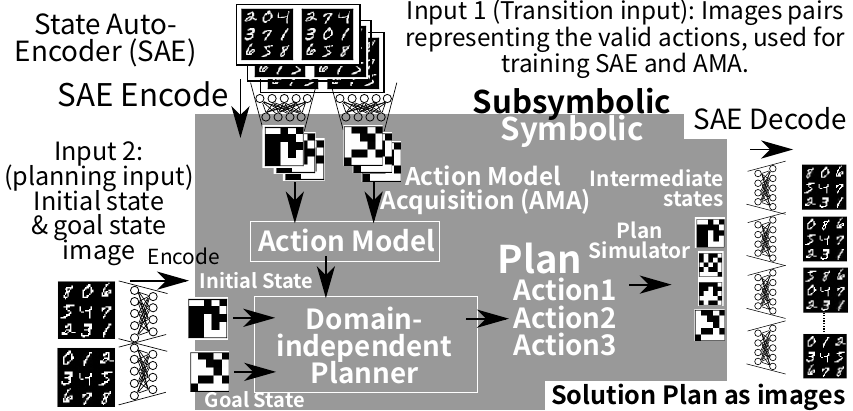}
 \caption{Classical planning in latent space:
It uses the learned State Autoencoder (\refig{sae}) to convert pairs of images $(\before,\after)$ to symbolic transitions, from which the AMA component generates an action model.
It also encodes the initial and goal state images into symbolic initial/goal states.
A classical planner finds the symbolic solution plan.
Finally, intermediate states in the plan are decoded back to a human-comprehensible image sequence.}
\label{fig:overview}
\end{figure}

\latentplanner takes two inputs.
The first input is the \emph{transition input} $Tr$, a set of pairs of raw data.
Each pair $tr_i=(\before_i, \after_i) \in Tr$ represents a transition of the environment before and after some action is executed.
The second input is the \emph{planning input} $(i, g)$, a pair of raw data, which corresponds to the initial and the goal state of the environment.
The output of \latentplanner is a data sequence representing the plan execution that reaches $g$ from $i$.
While the original paper used an image-based implementation (``data'' = raw images),
the type of data is arbitrary as long as it is compatible with neural networks.

\begin{figure}[tb]
 \includegraphics[width=\linewidth]{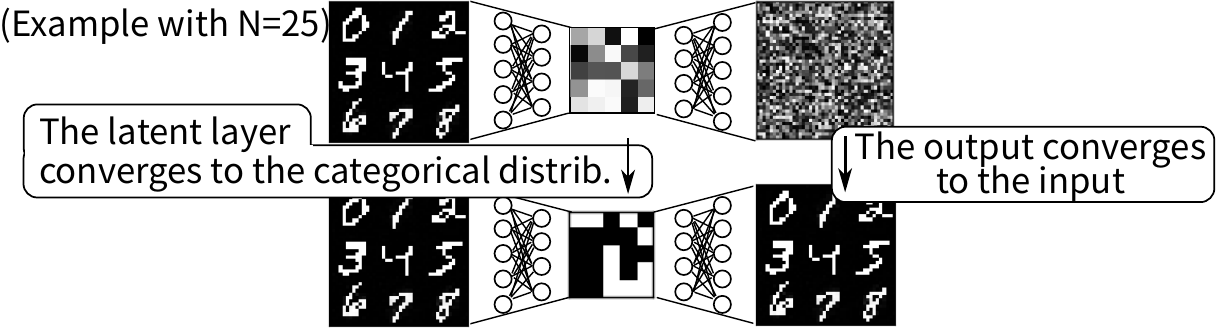}
 \caption{State AutoEncoder, a
 Variational AutoEncoder \cite{kingma2014semi} using Gumbel-Softmax \cite{jang2016categorical} reparametrization in its
 latent layer.}
 \label{sae}
\end{figure}

\latentplanner works in 3 phases.
In Phase 1, a \emph{State Autoencoder} (SAE) (\refig{sae}) learns a bidirectional mapping between raw data (e.g., images)
 and propositional states from a set of unlabeled, random snapshots of the environment, in an unsupervised manner.
The $Encode$ function maps images to propositional states, and $Decode$ function maps the propositional states back to images.
After training the SAE from $\braces{\before_i, \after_i\ldots}$,
it applies $Encode$ to each $tr_i \in Tr$ and obtain $(Encode(\before_i),$ $Encode(\after_i))=$ $(s_i,t_i)=$ $\overline{tr}_i\in \overline{Tr}$,
the symbolic representations (latent space vectors) of the transitions.

In Phase 2, an Action Model Acquisition (AMA) method learns an action model from $\overline{Tr}$ in an unsupervised manner.
The original paper proposed two approaches: AMA$_1$ is an oracular model which directly generates a PDDL without learning,
and AMA$_2$ is a neural model that approximate AMA$_1$ by learning from examples.

In Phase 3, a planning problem instance is generated from the planning input $(i,g)$.
These are converted to symbolic states by the SAE, and the symbolic planner solves the problem.
For example, an 8-puzzle problem instance consists of an image of the start (scrambled) configuration of the puzzle ($i$), and an image of the solved state ($g$).

Since the intermediate states comprising the plan are SAE-generated latent bit vectors, the ``meaning'' of each state (and thus the plan) is not clear to a human observer.
However, in the final step, \latentplanner obtains a step-by-step, human-comprehensive visualization of the plan execution
by $Decode$'ing the latent bit vectors for each intermediate state.
This is the reason the bidirectionality of the mapping is required in this framework.

The key concept of the SAE in \latentplanner is the use of \textbf{Gumbel-Softmax} \cite[GS]{jang2016categorical} activation function.
The output of GS (latent representation) converges to a 1-hot vector of $M$ categories.
This allows the SAE to obtain a discrete binary representation by setting $M$=2
and \latentplanner uses it for classical planning.
The output of a single Gumbel-Softmax unit $GS(\boldsymbol{\pi}) = \vz \in [0,1]^M$ is a one-hot vector representing $M$ categories, e.g.,
when $M=2$ the categories can be seen as $\braces{\text{false},\text{true}}$ and $\vz=\parens{0,1}$ represents ``true''.
(Note: There is no explicit meaning assigned to each category.)
The input $\boldsymbol{\pi} \in [0,1]^M$ is a class probability vector, e.g., $\parens{.2,.8}$.
Gumbel-Softmax is derived from Gumbel-Max technique \cite[\refeq{eq:gumbelmax}]{maddison2014sampling}
for drawing a categorical sample 
\begin{align}
z_i &= \textbf{if}\ (i \ \text{is}\ \arg \max_i (g_i+\log \pi_i)) \ \textbf{then}\ 1\ \textbf{else}\ 0 \label{eq:gumbelmax} \\
z_i &= \text{Softmax}((g_i+\log \pi_i)/\tau).                   \label{eq:gumbelsoftmax}
\end{align}
from $\boldsymbol{\pi}$
where $g_i$ is a sample drawn from
 $\text{Gumbel}(0,1) =-\log (-\log u)$ where $u=\text{Uniform}(0,1)$ \cite{gumbel1954statistical}.
Gumbel-Softmax approximates the argmax with a Softmax to make it differentiable (\refeq{eq:gumbelsoftmax}).
``Temperature'' $\tau$ controls the magnitude of approximation, which is annealed to 0 by a certain schedule.
The output $\vz$ converges to a discrete one-hot vector when $\tau\rightarrow 0$.

\begin{figure}[tb]
 \centering
 \includegraphics[width=0.8\linewidth]{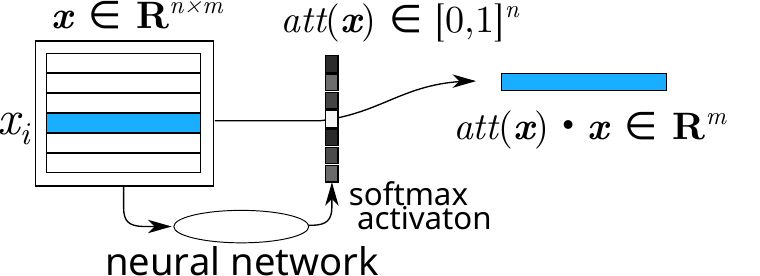}
 \caption{An attention mechanism selecting an $m$-dimensional vector out of $n$ vectors.}
 \label{attention}
\end{figure}

\textbf{Attention} is a recent mechanism that enhances the performance of
neural network for various cognitive tasks including machine translation \cite{bahdanau2017neural},
object recognition \cite{mnih2014recurrent},
image captioning \cite{xu2015show},
and Neural Turing Machine \cite{neuraltm}.
Its fundamental idea is to learn an attention function that extracts a single element from multiple elements.
The function is represented as a neural network and is trained unsupervised.
Typically, the output is activated by a Softmax, which normalizes the sum to 1.
The function can be formulated as
$\textit{att}(\vx)=\va$, where $\vx \in \R^{n\times m}$ is a set of $n$ elements $\vx = (\vx_1, \ldots \vx_n)$, $\vx_i \in \R^{m}$
and $\va=(a_1\ldots a_n) \in [0,1]^n$ satisfies $\sum_i a_i = 1$.
It can extract an element $\vx_i$ of $\vx$ using a dot product $\va \cdot \vx$.
For example, if $\vx=((2,0),(3,3),(1,2))$
and $\va=\parens{0,0.9,0.1}$, $\va \cdot \vx = (2.8, 2.9)$, which almost extracts $(3,3)$.

\section{High-Level Overview}
\label{requirements}

In order to find a first-order logic representation of the environment from raw data,
we perform the following processes (\refig{predicate-symbol-grounding}):
(1) \emph{Object detection}
identifies and extracts a set of regions from the raw data that contain objects.
(2) \emph{Predicate symbol grounding} (PSG) finds
the boolean functions that take several object feature vectors as the arguments.

While both processes are nontrivial, there are significant
advances in (1) recently.
Object recognition in computer vision \cite[YOLO]{redmon2016you} or
named entity (noun / ``objects'') recognition \cite{nadeau2007survey,mohit2014named} in Natural Language Processing,
are both becoming increasingly successful.
In this paper, therefore, we do not address (1)
and use a dataset that is already segmented into image patches and bounding boxes.
In principle, we could extract the object vectors with these external systems.

Next, PSG identifies a finite set of boolean functions
(predicates) from the feature input, by
learning to select the argument list from the input and
detecting the common patterns between the objects that define a relation.
As a result, we obtain the first-order logic representation of the input
as a list of FOL statements such as \textit{pred$_2$(obj$_1$, obj$_2$)=true},
where the system automatically learns to extract the arguments from the inputs,
and also decides the semantics of the predicates by itself, in an unsupervised manner.

While some might worry about the interpretability of the predicates with unknown semantics
and its compatibility to the existing knowledge base based on human-made symbols,
full autonomy is a valuable option that is orthogonal to the interpretability.
While interpretability is essential in the normal operations of, e.g., space exploration applications,
autonomy would be critical in an emergency situation. If a system lost contact to the human operators
in a unknown environment, it must learn a new representation online
(Imagine being abducted by an alien spaceship with shiny walls everywhere!
 A more realistic example is falling into an underground cave).
Moreover, a typical knowledge base is incompatible with raw observations such as images.

\section{First-Order State AutoEncoder}
\label{fosae}

\begin{figure*}[tb]
 \centering
 \includegraphics{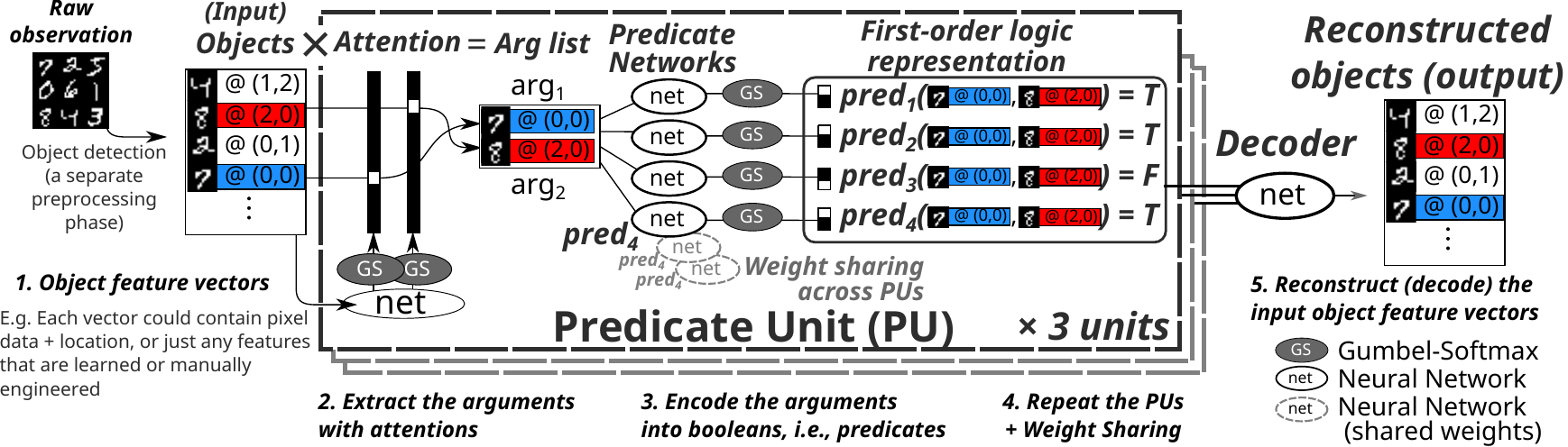}
 \caption{A First-Order State AutoEncoder (FOSAE) with $P=4$ predicates, arity $A=2$, and $U=3$ Predicate Units.
In this example, a feature vector consists of the pixel values and the $(x,y)$ location of an 8-Puzzle tile.}
 \label{fig:fosae}
\end{figure*}

We now introduce the core contribution of this paper,
First-Order State AutoEncoder (FOSAE, \refig{fig:fosae}), a
neural architecture which performs PSG and obtains a representation
compatible with symbolic reasoning systems such as classical planners.

\textbf{(\refig{fig:fosae}, 1)} Overall, the system follows the autoencoder architecture that takes
feature vectors of multiple objects in the environment as the input and reconstructs them as the output.
The form of the feature vector for each object is entirely problem/environment dependent:
It could be a hand-crafted feature vector,
a flattened vector of the raw pixel values for the object,
or a latent space vector automatically generated from the image array by an additional feature learning system (such as an autoencoder).
Let $\vx_n \in \R^F$ be a $F$-dimensional feature vector representing each object and
$\vx = (\vx_1, \ldots \vx_N)\in \R^{N\times F}$ be the input matrix representing the set of $N$ objects.

FOSAE consists of multiple instances of \emph{Predicate Unit},
a unit that (1) learns to extract an argument list from the input and
(2) computes the boolean values of the predicates given the extracted argument list.
The number of units $U$, the arity of predicates $A$ and the number of predicates $P$ are hyperparameters
which should be sufficiently large so that the network can encode enough information into a boolean vector
and then reconstruct the input.
If the network does not converge into a sufficiently low reconstruction loss,
we can increase these parameters until it does.
How to run this iteration efficiently is a hyperparameter tuning problem which is out of the scope of this paper.

\textbf{(\refig{fig:fosae}, 2)} In order to extract the arguments of the predicates, we use multiple attention networks (\refsec{background}).
The use of attention avoids enumerating $O(N^A)$ object tuples for $N$ objects as was done in the previous work.
There are $A$ attentions in each PU, thus each PU extracts $A$ objects from the $N$ objects in the input.
With $U$ PUs, there are $U\times A$ attentions.

An attention network is implemented
as a 2 fully-connected networks ending with a Gumbel-Softmax activation.
Unlike previous work which uses a Softmax in the output,
where the attention vectors take the continuous probability values produced by Softmax,
we instead use Gumbel-Softmax which converges to a discrete one-hot
vector so that the meaning of the extracted objects are clear.
For example, if an attention vector for an argument takes a value $(0,1,0)$, it
is clearly extracting the 2nd object in 3 objects, while if it were $(0,0.5,0.5)$,
it is unclear what was selected.

To extract the arguments, we take a dot-product (\refsec{background}) of
$\vx \in \R^{N\times F}$ and the $U\times A$ attention vectors
$\mathbf{att}(\vx) \in \R^{U\times A\times N}$,
where $\textit{att}_{ua}(\vx) \in [0,1]^N, 1\leq u \leq U, 1\leq a \leq A$.
This results in $U$ sets of $A$ arguments:
$\mathbf{att}(\vx) \cdot \vx = \vg \in \R^{U\times A \times F}$.
For example, $\vg_2$ can be seen as the argument list for the second PU,
and $\vg_{2,3}$ as its third argument.

\textbf{(\refig{fig:fosae}, 3)} Next, in each $u$-th PU, a set of NNs called \emph{Predicate Network} (PN) using
Gumbel-Softmax takes the arguments $\vg_u = (\vg_{u1} \ldots \vg_{uA})$ and
outputs a discrete 1-hot vector of 2 categories, which means true if the
first cell is 1, and false otherwise.
There are $P$ PNs where each PN $\textit{pred}_p$ ($1\leq p \leq P$) returns a single boolean value
and models a first order predicate $\textit{pred}_p(\vg_{u1} \ldots \vg_{uA}) \in \braces{0,1}$.
The boolean values have the same role as the representation discovered by the propositional SAE.

\textbf{(\refig{fig:fosae}, 4)} Attentions and PNs form a single PU.
We repeat such PUs $U$ times, which results in $U\times P$ total propositions.
While the weights in the attention functions ($\textit{att}_{ua}$) are specific to each PU,
the PN weights for $\textit{pred}_p$ are shared across PUs (hence it lacks the subscript $u$ here).
This makes the boolean function $\textit{pred}_p$ in different PUs identical to each other,
and force them to learn common relations among the different arguments
because PNs take different arguments in each PU.
We implemented PNs as 2-layer fully connected networks, but this is up to hyperparameter tuning.

\textbf{(\refig{fig:fosae}, 5)} Finally, the input object vectors are reconstructed from the propositional representation
by concatenating the boolean outputs from all PUs and feeding them to the decoder.
The requirement that the decoder should reconstruct the input is acting as a constraint:
In order to reconstruct the input,
the attentions should cover a sufficiently diverse set of objects
and also the different predicates should carry significantly different meanings.
This avoids the mode collapse of the attentions and the predicates.
The network is optimized by the Adam \cite{kingma2014adam} and
backpropagation with the square error loss.

\section{Modeling 8-Puzzle Instances}
\label{experiments-8puzzle}

\begin{figure}[tb]
 \centering
 \includegraphics[width=0.7\linewidth]{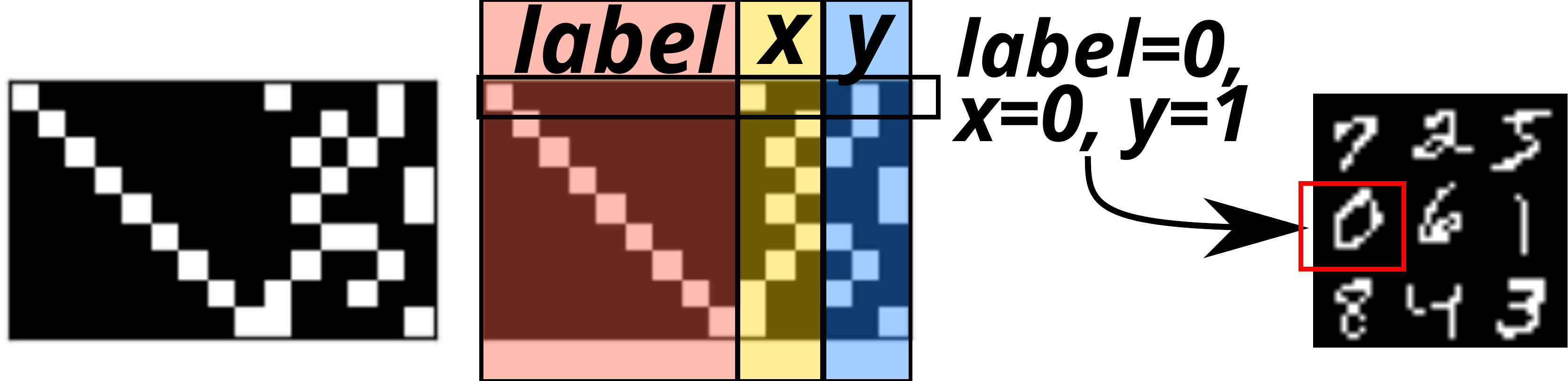}
 \caption{
A single 8-puzzle state as a 9x15 matrix, representing 9 objects of 15 features.
The first 9 features are the tile numbers
and the other 6 features are the 1-hot x/y-coordinates.
}
 \label{8puzzle-features}
\end{figure}

In order to evaluate FOSAE, 
we created a toy environment
of 8-puzzle states using the feature vectors shown in \refig{8puzzle-features}.
The purpose of this experiment is to show the feasibility of the first order predicate symbols
discovered by FOSAE
as the source of a PDDL planning model as well as
to show the evidence that the relational model is indeed necessary for modeling a complex environment.
Each feature vector as an object consists of 15 features, 9 of which represent the tile number (object ID) and the remaining 6
represent the coordinates.
Each data point has 9 such vectors, corresponding to the 9 objects in a single tile configuration.
We generated 20000 transition inputs (state pairs) which are divided into
18000 (training set) and 2000 (test set).

\begin{figure*}[htb]
 \centering
 \includegraphics[width=\linewidth]{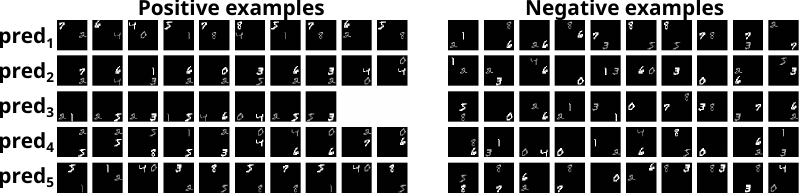}
 \caption{
 The positive/negative examples of the arguments for the first 5
 predicates of $(U,A,P)=(25,2,50)$. The first/second argument is
 visualized in white / gray.
We could interpret the condition for each $\textit{pred}_p$ returning true as follows:
 $\textit{pred}_1$: $(x_1,y_1)=(0,0)$ and $(x_2,y_2)=(*,1)$.
 $\textit{pred}_2$: $(x_1,y_1)=(2,1)$ and $(x_2,y_2)=(2,*)$.
 $\textit{pred}_3$: $y_1=y_2=2$ and $x_1 < x_2$.
 $\textit{pred}_4$: $(x_1,y_1)=(2,*)$ and $(x_2,y_2)=(2,0)$.
 $\textit{pred}_5$: $(x_1,y_1)=(1,0)$ and $(x_2,y_2)=(2,*)$.
$*$ denotes the wildcard.
 }
 \label{8puzzle-argument-visualization}
 \label{8puzzle-decision-tree}
\end{figure*}

\subsubsection{Are Higher Arity Predicates Truly Necessary?}

Previous work on relational structures has not yet provided evidence that
they help to model the environment and extract abstract knowledge.
For example,
it is possible that
even if
a relational structure like RN \cite{santoro2017simple} extracts multiple arguments,
the succeeding layers may ignore some arguments by assigning zero weights,
essentially modeling just unary predicates (i.e. attributes)
rather than the structural relationships.
We need to show direct evidence that the FOSAE extracts the \emph{essential} higher-arity relations,
without entangling the system with the policy learning structures.

One way to show that PNs are extracting higher-arity relations is to compare the minimum capacity
required for the network to reconstruct the input for each arity $A$.
The intuition here is that high-arity relations provide abstract knowledge that helps to compress the information.
For example, with a \texttt{(x-next ?o1 ?o2)} relationship,
the network does not have to encode the absolute information for every objects (e.g., \texttt{(x0 o1) (y0 o1) (x1 o2) (y0 o2)}: 4bits)
and rather the minimal amount of absolute information tied with a couple of relative information 
(e.g., \texttt{(x0 o1) (y0 o1) (x-next o1 o2)}: 3bits).
This is in line with the concepts in generic compression algorithms
which try to minimize the redundant information.
Note that we do not claim that the meaning of the predicates extracted by a PN is always such adjacency relations.
In our unsupervised setting, we do not control (supervise) the
type of relations the network decides to represent.

\begin{figure}[tb]
 \centering
 \includegraphics[width=0.32\linewidth]{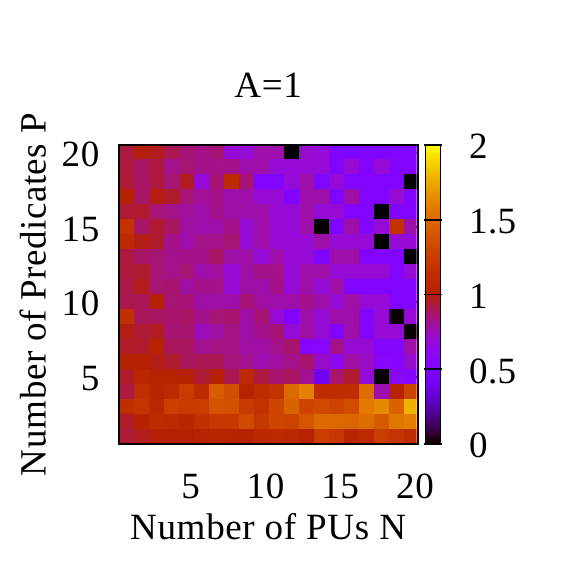}
 \includegraphics[width=0.32\linewidth]{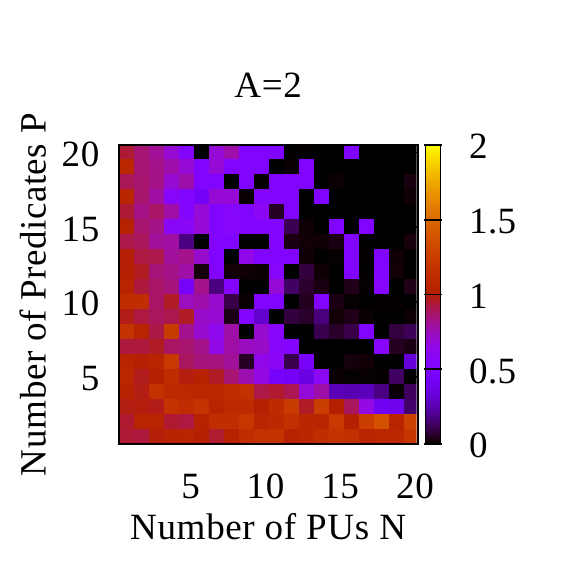}
 \includegraphics[width=0.32\linewidth]{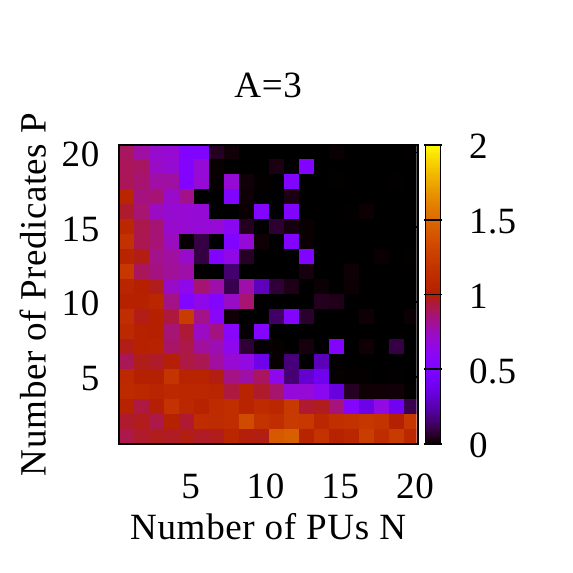}
 \caption{Contour plots of the reconstruction error of the test set
for $A$=1,2,3, $(U,P)\in [1..20]^2$.
It shows that the larger arity helps to learn the compact representation.}
 \label{arity-contour}
\end{figure}

We made the contour plots (\refig{arity-contour}) of the reconstruction errors
for the test set
with various $U,P,A$, and compared their Pareto fronts.
For the same $(U,P)$ pair, the size of the bottleneck layer (propositional vector) is $U\times P$ regardless of $A$, which makes
the direct comparison between different $A$ feasible.
We see that the arity plays a critical role in finding the more compact information,
demonstrating that structural relations contribute to building an abstract representation.

We also compared the number of trainable parameters (weights) in the network
because, for the same $(U,P)$, the larger arity means the larger number of parameters in the networks
which may help the training.
\reftbl{arity-parameters} shows the models with the fewest parameters among those achieved the reconstruction error $\leq 0.1$ for each $A$.
FOSAE with a larger arity can indeed be trained with fewer weights.

\begin{table}[tb]
 \centering
 \begin{tabular}{ccccc}
  $A$             & $U$      & $P$ & Propositions & Trainable parameters \\\hline
  1               & 18       & 5   & 90          & 287343               \\
  2               & 9        & 6   & \textbf{54} & \textbf{268273}      \\
  3               & 9        & 7   & 63          &  303302              \\
  9               & 1        & 171 & 171         & 811828               \\\hline
  \multicolumn{3}{l}{SAE (Asai 2018)}  & 18      & 3404467              \\\hline
 \end{tabular}
 \caption{
Configurations $(U,P)\in [1,20]^2$ for each $A$ that achieved the reconstruction error $\leq 0.1$
with the smallest trainable parameters.
Arity $A$=2,3
achieved the human-indiscernible accuracy with a fewer number of parameters than $A$=1,
while finding the smaller representations.
This shows that the relational structures indeed help to model
the environment by introducing a higher level of abstraction.
Next, with $A$=9,$U$=1 tested over $P\in[1,400]$,
the network is allowed to look at all objects and finds independent propositions
while sharing the FOSAE architecture and other hyperparameters.
It requires significantly larger parameters and latent representations.
Finally, while the standard SAE \cite{Asai2018} finds a more compressed representation,
it consumes 10x more weight parameters.
}
 \label{arity-parameters}
\end{table}

We also compared the combination $(A,U)=(9,1)$ with $P\in[1..400]$,
where 400 is the same maximum latent space capacity as the previous experiment.
Since the predicates are allowed to see all objects ($A$=9),
they are functions from the environment itself to a single boolean value,
i.e. propositions,
while maintaining the same FOSAE architecture.
\reftbl{arity-parameters} shows that this configuration performs poorly,
providing further evidence that the relations help
obtain a compact, abstract representation.

We also confirmed that the standard SAE \cite{Asai2018} consumes 10x more weight parameters
even when tuned to have the minimal number of propositions under the reconstruction error $\leq 0.1$.
This is because, in a fully propositional SAE, each proposition is independently learned
even if some propositions are carrying similar information for the different objects.
This is similar to what a convolutional layer is to a fully connected layer for image processing,
where the former uses the shared filters to process the different local image patches.

\subsubsection{Interpreting Predicates}

Next, we show how the hard attentions make the predicates interpretable through visualization.
In principle, we can visualize the objects in the images selected by the attentions
(e.g., monkeys, bananas in \refig{predicate-symbol-grounding})
using a decoder function that reconstructs the regions from feature vectors.
For the 8-puzzle feature vectors that we manually created for this experiment (\refig{8puzzle-features}),
we instead use a hand-crafted decoder that pastes the corresponding image patch for the tile
to the region specified by the xy-coordinates.
Thanks to the hard, discrete attention activated by Gumbel-Softmax, no two objects are mixed.

\refig{8puzzle-argument-visualization} shows the visualizations of the arguments given to the predicate networks
under hyperparameter $(U,A,P)=(25,2,50)$.
Each subfigure is a visualization of an argument list vector $\vg_u=(\vg_{u1}, \vg_{u2})$ randomly sampled from the dataset.
Examples in the same row correspond to one predicate,
where the left half represent the arguments which made the predicate true,
and the right half represents those which made it false.
We humans could recognize the patterns that are shared on the left-hand side of each row,
giving us the possibility of interpretation which is not available in the fully propositional representation.

\section{Evaluating Classical Planning Capability}

We show that the FOL representation generated by FOSAE is
a feasible and sound representation for classical planning.

We tested the FOSAE-generated representation with AMA$_1$ PDDL generator
and the Fast Downward \cite{Helmert04} classical planner.
AMA$_1$ is an oracular method
that takes the entire raw state transitions, encode each $\braces{\before_i, \after_i}$ pair with the SAE,
then instantiate each encoded pair into a grounded action schema.
It models the ground truth of the transition rules,
thus is useful for verifying the state representation.
Planning fails when SAE fails to encode a given init/goal image into
a propositional state that exactly matches one of the search nodes.
While there are several learning-based AMA methods that approximate AMA$_1$
(e.g., AMA$_2$ \cite{Asai2018} and Action Learner \cite{amado2018goal}),
there is information loss between the learned action model and the original search space generated by FOSAE,
which make them unsuitable for our purpose of testing the feasibility of the representation.

AMA$_1$ is a fully propositional AMA method.
To run AMA$_1$, we use the propositional output of PNs as the state representation,
not the first-order representation.
We leave the task of obtaining the lifted action model as future work.

We invoke Fast Downward with blind heuristics in order to remove the effect of the heuristics.
This is primarily because AMA$_1$ generates a huge PDDL model containing every transition
which results in an excessive runtime for initializing any sophisticated heuristics.
The scalability issue caused by using a blind heuristics is not an issue
since the focus of this evaluation is on the feasibility of the representation.

\subsection{8 Puzzle}

We generated 40 instances of 8-puzzle each generated by a random walk from
the goal state. 40 instances consist of 20 instances each generated by a 7-steps random walk
and another 20 by 14 steps.
Results (\reftbl{8puzzle-ama1}) show that the planner successfully solves all problems,
demonstrating that the representation grounded by the FOSAE is sound for planning.

\begin{table}[tb]
 \centering
 \begin{tabular}{|c|c|c|c|}
\hline
random walk & \#solved & search time [sec]   & cost           \\
steps    & / \#total   &  (mean) & (mean) \\\hline
\multicolumn{4}{|c|}{\small 8 Puzzle} \\\hline
7 steps  & 20/20 & 0.0014  & 7.00   \\
14 steps & 20/20 & 0.0236  & 13.60  \\\hline
\multicolumn{4}{|c|}{\small 3-blocks Blocksworld} \\\hline
3 steps     & 10/10       & 0.0013            & 2.2    \\
7 steps     & 10/10       & 0.0062            & 3.6    \\
14 steps    & 10/10       & 0.0226            & 5.7    \\\hline
 \end{tabular}
 \caption{The number of instances solved by FOSAE + AMA$_1$ on
the object vector based 8-puzzle instances and
3-blocks Blocksworld instances.
The number of random walk steps counts the steps to generate a goal state from the initial state.
The cost column in the 14 steps instances shows that it finds a solution shorter than 14,
which is natural because the 14 steps random walk
do not always correspond to the optimal solution of the puzzle.}
 \label{8puzzle-ama1}
 \label{blocks-ama1}
\end{table}

\begin{figure}[tb]
 \centering
 \includegraphics[width=\linewidth]{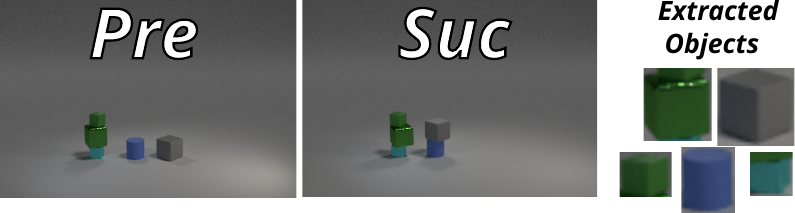}
 \caption{
An example Blocksworld transition.
Each state has a perturbation from the jitter in the light positions and the ray-tracing noise.
Other objects may intrude the extracted regions.
Objects have different sizes, colors, shapes (cube or cylinder) and surface materials (metal or rubber).
}
 \label{blocks-example}
\end{figure}

\begin{figure*}[tb]
 \centering
 \resizebox{.95\textwidth}{!}{
 \begin{minipage}[b]{0.32\linewidth}
  \centering
 \includegraphics[width=0.46\linewidth]{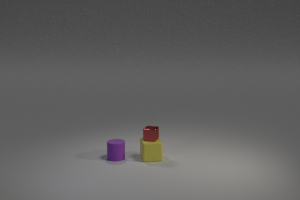}
 \includegraphics[width=0.46\linewidth]{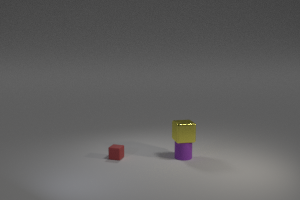}\\
 \end{minipage}
 \begin{minipage}[b]{0.23\linewidth}
  \centering
 \includegraphics[width=0.31\linewidth]{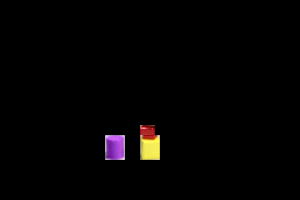}
 \includegraphics[width=0.31\linewidth]{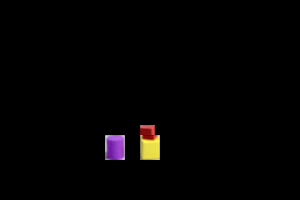}
 \includegraphics[width=0.31\linewidth]{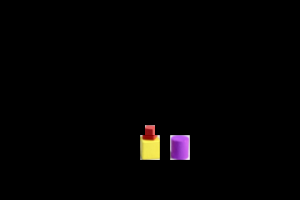}\\
 \includegraphics[width=0.31\linewidth]{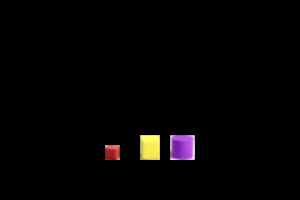}
 \includegraphics[width=0.31\linewidth]{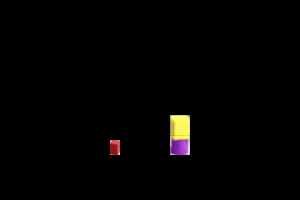}
 \includegraphics[width=0.31\linewidth]{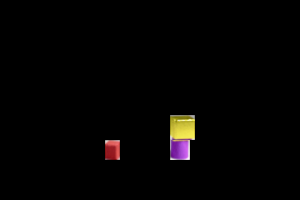}
 \end{minipage}
 }
 \caption{
(\textbf{middle}) The initial/goal state of a Blocksworld instance.
(\textbf{right}) The solution to this problem reconstructed from the latent vector.
It unpolishes the red cube, then moves the cylinder, the red cube, the yellow cube and then polishes the yellow cube.
The colors of blocks are affected by the preprocessing (histogram normalization)
which helps the training.
}
 \label{blocks-example}
 \label{blocks-results-visualized}
\end{figure*}

\begin{figure}[htb]
 \includegraphics[width=0.24\linewidth]{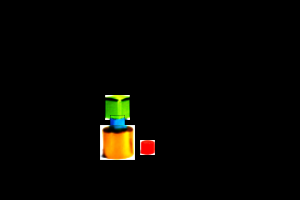}
 \includegraphics[width=0.24\linewidth]{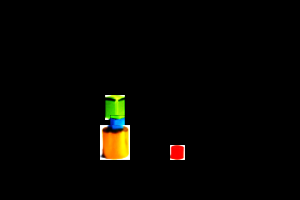}
 \includegraphics[width=0.24\linewidth]{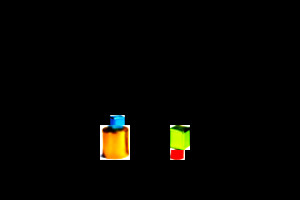}
 \includegraphics[width=0.24\linewidth]{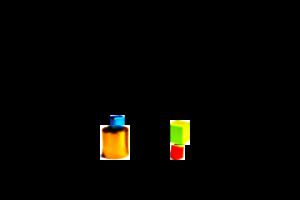}
 \caption{A successful plan execution for 4 blocks.}
 \label{blocksworld-success}
\end{figure}

\subsection{Photo-Realistic Blocksworld}
\label{experiments-blocks}

\begin{figure}[tb]
  \centering
 \includegraphics[width=0.6\linewidth]{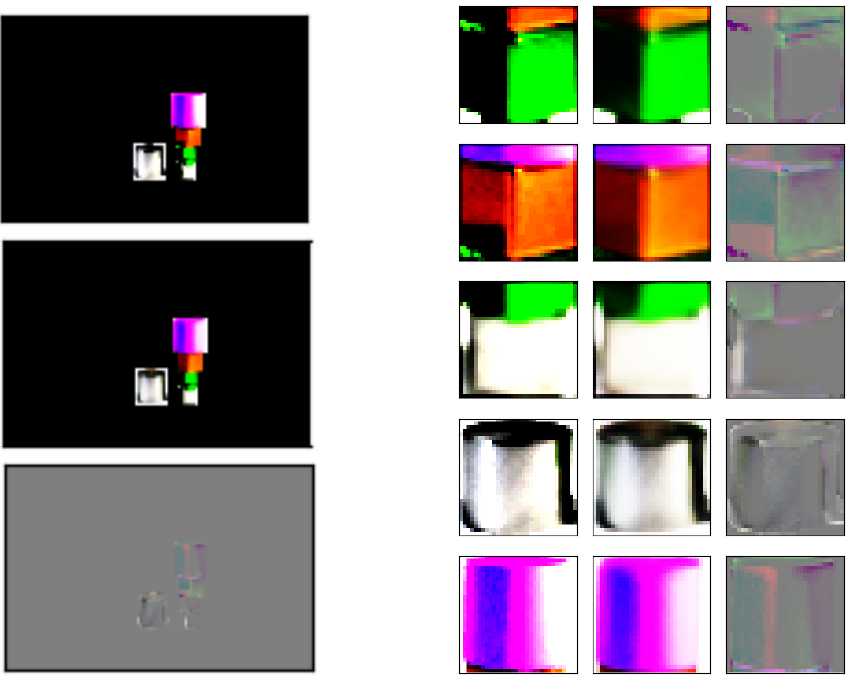}
 \caption{
(\textbf{left})
The visualization of the autoencoding result using FOSAE in a 5 blocks, 3 stacks environment.
The input (ground truth) is presented on the top, and the output (reconstruction) is presented in the middle.
The bottom figure shows the difference between the input and the output.
(\textbf{right})
The enlarged results of the reconstructions for the individual image patches.
From the left, the ground truth, the reconstruction, and the pixel value difference.
}
 \label{blocks-autoencoding}
\end{figure}

In order to test the ability of the system in a more realistic
environment, we prepared a photo-realistic Blocksworld dataset
(\refig{blocks-example}) \cite{Asai2018b}
which contains the blocks world states rendered by Blender 3D engine.
There are several cylinders or cubes of
various colors and sizes and two surface materials (Metal/Rubber)
stacked on the floor, just like in the usual STRIPS Blocksworld domain.
In this domain, three actions are performed:
\texttt{move} a block onto another stack or on the floor,
and \texttt{polish}/\texttt{unpolish} a block, i.e., change the
surface of a block from Metal to Rubber or vice versa.
All actions are applicable only when the block is on top of a stack or on the floor.
The latter actions allow changes in the non-coordinate features of object vectors.

The dataset generator produces a 300x200 RGB image and a state description which contains
the bounding boxes (bbox) of the objects.
Extracting these bboxes is an object recognition task we do not address in this paper,
and ideally, should be performed by a system like YOLO \cite{redmon2016you}.
We resized the extracted image patches in the bboxes to 32x32 RGB, flattened it into a 3072-D vector,
and concatenated it with the bbox vector.
The bbox vector is 200-dimensional and is generated by
discretizing $(x_1,y_1,x_2,y_2)$ by 5 pixels and encoding it as a 1-hot vector (60/40 categories for each $x$/$y$-axis),
resulting in 3072+200=3272 features per object.
FOSAE encodes and reconstructs a set of feature vectors, each containing both the pixel and the bbox information.
The feature vectors can be visualized by pasting the pixels into the bbox on a black canvas.

The generator enumerates all possible states/transitions
(480/2592 for 3 blocks and 3 stacks;
 5760/34560 for 4 blocks and 3 stacks;
 80640/518400 for 5 blocks and 3 stacks).
For training the FOSAE,
we used 432 (90\%, 3 blocks 3 stacks), 2500 (4 blocks 3 stacks), and 4500 states (5 blocks, 3 stacks), respectively.
We chose $(U,A,P)=(10,2,100)$ as the hyperparameter.
An example reconstruction is shown in \refig{blocks-autoencoding}.

We then solved 30 planning instances generated by
taking a random initial state and choosing the goal states by the 3, 7, or 14 steps random walks (10 instances each).
The system correctly solved all instances,
where the correctness of the plan was checked manually.
\refig{blocks-results-visualized} shows an example solution generated from
the intermediate states of the plan.
We only performed planning for the 3 blocks environment 
due to the sheer size of the PDDL model generated by AMA$_1$ which contains
518400 actions and required more than 128GB memory to preprocess the model into a SAS+ format.
We later performed some 4-blocks experiments and obtained success (\refig{blocksworld-success}).

\section{Discussion and Conclusion}
\label{conclusion}

We proposed First-Order State AutoEncoder, 
a neural architecture which
grounds/extracts first-order logical
predicates from the environment without human supervision.
Unlike any existing work to our knowledge,
the training is fully automated (no manual tagging / no predefined reinforcement signals)
and the resulting representation is
interpretable, verifiable and compatible with symbolic systems
such as classical planners.
 
We also provided the first empirical evidence that the relations between
the objects help to model the environment by 
testing the architecture with various predicate arities.
FOSAE exclusively models the environment,
unlike a black-box mixture of the policy and the representation
learned by Reinforcement Learning frameworks.

While the predicates grounded by the FOSAE are anonymous symbols
that do not necessarily correspond to the human symbols (e.g., \texttt{next}),
an interesting avenue of future work is to adapt the system to the semi-/supervised setting,
where the supervised signals are fed into the latent layer \cite{kingma2014semi}.
This could teach FOSAE the human symbols in the hand-coded knowledge base.
Since semi-supervised/supervised learning are in general considered easier
than unsupervised learning (done by our FOSAE), this is a promising direction.

Finally, note that we did not address the full FOL generalization, that is,
the learned FOL statements are quantifier-free, grounded representation.
Despite the limitation, this work is an important step toward the full FOL generalization
-- after all, to lift a FOL formula, the formula must be generated from the subsymbolic input in the first place.
The next step is to lift the representation in order to
extract action rules, axioms, and invariants,
which enable the sophisticated heuristics developed in the heuristic search literature.
Extending or leveraging the existing Action Model Acquisition methods
(e.g., AMA$_1$, AMA$_2$, Action Learner \cite{amado2018goal}, \cite{YangWJ07,MouraoZPS12,CresswellMW13})
is an important avenue for future work.

\fontsize{9pt}{10pt}
\selectfont

\bibliographystyle{aaai}

\begin{thebibliography}{}

\bibitem[\protect\citeauthoryear{Amado \bgroup et al\mbox.\egroup
  }{2018}]{amado2018goal}
Amado, L.; Pereira, R.~F.; Aires, J.; Magnaguagno, M.; Granada, R.; and
  Meneguzzi, F.
\newblock 2018.
\newblock {Goal Recognition in Latent Space}.
\newblock In {\em {Proceedings of the International Joint Conference on Neural
  Networks (IJCNN)}}.

\bibitem[\protect\citeauthoryear{Asai and Fukunaga}{2018}]{Asai2018}
Asai, M., and Fukunaga, A.
\newblock 2018.
\newblock {Classical Planning in Deep Latent Space: Bridging the
  Subsymbolic-Symbolic Boundary}.
\newblock In {\em {Proceedings of AAAI Conference on Artificial Intelligence}}.

\bibitem[\protect\citeauthoryear{Asai}{2018}]{Asai2018b}
Asai, M.
\newblock 2018.
\newblock Photo-realistic blocksworld dataset.
\newblock {\em arXiv:1812.01818}.

\bibitem[\protect\citeauthoryear{Bahdanau, Cho, and
  Bengio}{2015}]{bahdanau2017neural}
Bahdanau, D.; Cho, K.; and Bengio, Y.
\newblock 2015.
\newblock {Neural Machine Translation by Jointly Learning to Align and
  Translate}.
\newblock In {\em {Proceedings of the International Conference on Learning
  Representations}}.

\bibitem[\protect\citeauthoryear{Battaglia \bgroup et al\mbox.\egroup
  }{2018}]{battaglia2018relational}
Battaglia, P.~W.; Hamrick, J.~B.; Bapst, V.; Sanchez-Gonzalez, A.; Zambaldi,
  V.; Malinowski, M.; Tacchetti, A.; Raposo, D.; Santoro, A.; Faulkner, R.;
  et~al.
\newblock 2018.
\newblock {Relational inductive biases, deep learning, and graph networks}.
\newblock {\em arXiv:1806.01261}.

\bibitem[\protect\citeauthoryear{Cresswell, McCluskey, and
  West}{2013}]{CresswellMW13}
Cresswell, S.; McCluskey, T.~L.; and West, M.~M.
\newblock 2013.
\newblock Acquiring planning domain models using \emph{LOCM}.
\newblock {\em Knowledge Eng. Review} 28(2):195--213.

\bibitem[\protect\citeauthoryear{Garnelo, Arulkumaran, and
  Shanahan}{2016}]{garnelo2016towards}
Garnelo, M.; Arulkumaran, K.; and Shanahan, M.
\newblock 2016.
\newblock {Towards Deep Symbolic Reinforcement Learning}.
\newblock {\em arXiv preprint arXiv:1609.05518}.

\bibitem[\protect\citeauthoryear{Graves \bgroup et al\mbox.\egroup
  }{2016}]{neuraltm}
Graves, A.; Wayne, G.; Reynolds, M.; Harley, T.; Danihelka, I.;
  Grabska-Barwi{\'n}ska, A.; Colmenarejo, S.~G.; Grefenstette, E.; Ramalho, T.;
  Agapiou, J.; et~al.
\newblock 2016.
\newblock {Hybrid Computing using a Neural Network with Dynamic External
  Memory}.
\newblock {\em Nature} 538(7626):471--476.

\bibitem[\protect\citeauthoryear{Gumbel and
  Lieblein}{1954}]{gumbel1954statistical}
Gumbel, E.~J., and Lieblein, J.
\newblock 1954.
\newblock {Statistical theory of extreme values and some practical
  applications: A series of lectures}.

\bibitem[\protect\citeauthoryear{Harnad}{1990}]{harnad1990symbol}
Harnad, S.
\newblock 1990.
\newblock {The symbol grounding problem}.
\newblock {\em Physica D: Nonlinear Phenomena} 42(1-3):335--346.

\bibitem[\protect\citeauthoryear{Helmert}{2004}]{Helmert04}
Helmert, M.
\newblock 2004.
\newblock {A Planning Heuristic Based on Causal Graph Analysis}.
\newblock In {\em {Proceedings of the International Conference on Automated
  Planning and Scheduling(ICAPS)}},  161--170.

\bibitem[\protect\citeauthoryear{Hoffmann and Nebel}{2001}]{hoffmann01}
Hoffmann, J., and Nebel, B.
\newblock 2001.
\newblock {The FF Planning System: Fast Plan Generation through Heuristic
  Search}.
\newblock {\em {J. Artif. Intell. Res.(JAIR)}} 14:253--302.

\bibitem[\protect\citeauthoryear{Jang, Gu, and
  Poole}{2017}]{jang2016categorical}
Jang, E.; Gu, S.; and Poole, B.
\newblock 2017.
\newblock {Categorical Reparameterization with Gumbel-Softmax}.
\newblock In {\em {Proceedings of the International Conference on Learning
  Representations}}.

\bibitem[\protect\citeauthoryear{Kingma and Ba}{2015}]{kingma2014adam}
Kingma, D., and Ba, J.
\newblock 2015.
\newblock {Adam: A Method for Stochastic Optimization}.
\newblock In {\em {Proceedings of the International Conference on Learning
  Representations}}.

\bibitem[\protect\citeauthoryear{Kingma \bgroup et al\mbox.\egroup
  }{2014}]{kingma2014semi}
Kingma, D.~P.; Mohamed, S.; Rezende, D.~J.; and Welling, M.
\newblock 2014.
\newblock {Semi-Supervised Learning with Deep Generative Models}.
\newblock In {\em {Advances in Neural Information Processing Systems}},
  3581--3589.

\bibitem[\protect\citeauthoryear{LeCun, Bengio, and
  Hinton}{2015}]{lecun2015deep}
LeCun, Y.; Bengio, Y.; and Hinton, G.
\newblock 2015.
\newblock {Deep Learning}.
\newblock {\em Nature} 521(7553):436.

\bibitem[\protect\citeauthoryear{Maddison, Tarlow, and
  Minka}{2014}]{maddison2014sampling}
Maddison, C.~J.; Tarlow, D.; and Minka, T.
\newblock 2014.
\newblock {A* sampling}.
\newblock In {\em {Advances in Neural Information Processing Systems}},
  3086--3094.

\bibitem[\protect\citeauthoryear{Manhaeve \bgroup et al\mbox.\egroup
  }{2018}]{manhaeve2018deepproblog}
Manhaeve, R.; Duman{\v{c}}i{\'c}, S.; Kimmig, A.; Demeester, T.; and De~Raedt,
  L.
\newblock 2018.
\newblock {DeepProbLog: Neural Probabilistic Logic Programming}.
\newblock In {\em {Advances in Neural Information Processing Systems}}.

\bibitem[\protect\citeauthoryear{McDermott}{2000}]{McDermott00}
McDermott, D.~V.
\newblock 2000.
\newblock {The 1998 {AI} Planning Systems Competition}.
\newblock {\em {AI} Magazine} 21(2):35--55.

\bibitem[\protect\citeauthoryear{Mnih \bgroup et al\mbox.\egroup }{2015}]{dqn}
Mnih, V.; Kavukcuoglu, K.; Silver, D.; Rusu, A.~A.; Veness, J.; Bellemare,
  M.~G.; Graves, A.; Riedmiller, M.; Fidjeland, A.~K.; Ostrovski, G.; et~al.
\newblock 2015.
\newblock {Human-Level Control through Deep Reinforcement Learning}.
\newblock {\em Nature} 518(7540):529--533.

\bibitem[\protect\citeauthoryear{Mnih, Heess, and
  Graves}{2014}]{mnih2014recurrent}
Mnih, V.; Heess, N.; and Graves, A.
\newblock 2014.
\newblock Recurrent models of visual attention.
\newblock In {\em {Advances in Neural Information Processing Systems}},
  2204--2212.

\bibitem[\protect\citeauthoryear{Mohit}{2014}]{mohit2014named}
Mohit, B.
\newblock 2014.
\newblock {Named Entity Recognition}.
\newblock In {\em Natural language processing of semitic languages}. Springer.
\newblock  221--245.

\bibitem[\protect\citeauthoryear{Mour{\~a}o \bgroup et al\mbox.\egroup
  }{2012}]{MouraoZPS12}
Mour{\~a}o, K.; Zettlemoyer, L.~S.; Petrick, R. P.~A.; and Steedman, M.
\newblock 2012.
\newblock {Learning {STRIPS} Operators from Noisy and Incomplete Observations}.
\newblock In {\em {Proceedings of the International Conference on Uncertainty
  in Artificial Intelligence}},  614--623.

\bibitem[\protect\citeauthoryear{Nadeau and Sekine}{2007}]{nadeau2007survey}
Nadeau, D., and Sekine, S.
\newblock 2007.
\newblock {A Survey of Named Entity Recognition and Classification}.
\newblock {\em Lingvisticae Investigationes} 30(1):3--26.

\bibitem[\protect\citeauthoryear{Paccanaro and
  Hinton}{2001}]{paccanaro2001learning}
Paccanaro, A., and Hinton, G.~E.
\newblock 2001.
\newblock {Learning Distributed Representations of Concepts using Linear
  Relational Embedding}.
\newblock {\em IEEE Transactions on Knowledge and Data Engineering}
  13(2):232--244.

\bibitem[\protect\citeauthoryear{Redmon \bgroup et al\mbox.\egroup
  }{2016}]{redmon2016you}
Redmon, J.; Divvala, S.; Girshick, R.; and Farhadi, A.
\newblock 2016.
\newblock {You Only Look Once: Unified, Real-Time Object Detection}.
\newblock In {\em {Proceedings of IEEE Conference on Computer Vision and
  Pattern Recognition}},  779--788.

\bibitem[\protect\citeauthoryear{Rockt\"{a}schel and
  Riedel}{2017}]{rocktaschel2017end}
Rockt\"{a}schel, T., and Riedel, S.
\newblock 2017.
\newblock {End-to-end Differentiable Proving}.
\newblock In Guyon, I.; Luxburg, U.~V.; Bengio, S.; Wallach, H.; Fergus, R.;
  Vishwanathan, S.; and Garnett, R., eds., {\em {Advances in Neural Information
  Processing Systems}}. Curran Associates, Inc.
\newblock  3788--3800.

\bibitem[\protect\citeauthoryear{Russell \bgroup et al\mbox.\egroup
  }{1995}]{russell1995artificial}
Russell, S.~J.; Norvig, P.; Canny, J.~F.; Malik, J.~M.; and Edwards, D.~D.
\newblock 1995.
\newblock {\em {Artificial Intelligence: A Modern Approach}}, volume~2.
\newblock Prentice hall Englewood Cliffs.

\bibitem[\protect\citeauthoryear{Santoro \bgroup et al\mbox.\egroup
  }{2017}]{santoro2017simple}
Santoro, A.; Raposo, D.; Barrett, D.~G.; Malinowski, M.; Pascanu, R.;
  Battaglia, P.; and Lillicrap, T.
\newblock 2017.
\newblock A simple neural network module for relational reasoning.
\newblock In {\em {Advances in Neural Information Processing Systems}},
  4967--4976.

\bibitem[\protect\citeauthoryear{Scarselli \bgroup et al\mbox.\egroup
  }{2009}]{scarselli2009graph}
Scarselli, F.; Gori, M.; Tsoi, A.~C.; Hagenbuchner, M.; and Monfardini, G.
\newblock 2009.
\newblock {The Graph Neural Network Model}.
\newblock {\em IEEE Transactions on Neural Networks} 20(1):61--80.

\bibitem[\protect\citeauthoryear{Sourek \bgroup et al\mbox.\egroup
  }{2018}]{sourek2018lifted}
Sourek, G.; Aschenbrenner, V.; Zelezny, F.; Schockaert, S.; and Kuzelka, O.
\newblock 2018.
\newblock {Lifted Relational Neural Networks: Efficient Learning of Latent
  Relational Structures}.
\newblock {\em {J. Artif. Intell. Res.(JAIR)}} 62:69--100.

\bibitem[\protect\citeauthoryear{Steels}{2008}]{Steels2008}
Steels, L.
\newblock 2008.
\newblock {The Symbol Grounding Problem has been Solved. So What's Next?}
\newblock In de~Vega, M.; Glenberg, A.; and Graesser, A., eds., {\em {Symbols
  and Embodiment}}. Oxford University Press.

\bibitem[\protect\citeauthoryear{Taddeo and Floridi}{2005}]{taddeo2005solving}
Taddeo, M., and Floridi, L.
\newblock 2005.
\newblock {Solving the Symbol Grounding Problem: A Critical Review of Fifteen
  Years of Research}.
\newblock {\em Journal of Experimental \& Theoretical Artificial Intelligence}
  17(4):419--445.

\bibitem[\protect\citeauthoryear{van Steenkiste \bgroup et al\mbox.\egroup
  }{2018}]{van2018relational}
van Steenkiste, S.; Chang, M.; Greff, K.; and Schmidhuber, J.
\newblock 2018.
\newblock {Relational Neural Expectation Maximization: Unsupervised Discovery
  of Objects and their Interactions}.
\newblock In {\em {Proceedings of the International Conference on Learning
  Representations}}.

\bibitem[\protect\citeauthoryear{Xu \bgroup et al\mbox.\egroup
  }{2015}]{xu2015show}
Xu, K.; Ba, J.; Kiros, R.; Cho, K.; Courville, A.; Salakhudinov, R.; Zemel, R.;
  and Bengio, Y.
\newblock 2015.
\newblock Show, attend and tell: Neural image caption generation with visual
  attention.
\newblock In {\em {Proceedings of the International Conference on Machine
  Learning}},  2048--2057.

\bibitem[\protect\citeauthoryear{Yang, Wu, and Jiang}{2007}]{YangWJ07}
Yang, Q.; Wu, K.; and Jiang, Y.
\newblock 2007.
\newblock {Learning Action Models from Plan Examples using Weighted {MAX-SAT}}.
\newblock {\em {Artificial Intelligence}} 171(2-3):107--143.

\bibitem[\protect\citeauthoryear{Zambaldi \bgroup et al\mbox.\egroup
  }{2018}]{zambaldi2018relational}
Zambaldi, V.; Raposo, D.; Santoro, A.; Bapst, V.; Li, Y.; Babuschkin, I.;
  Tuyls, K.; Reichert, D.; Lillicrap, T.; Lockhart, E.; et~al.
\newblock 2018.
\newblock {Relational Deep Reinforcement Learning}.
\newblock {\em arXiv:1806.01830}.

\end{thebibliography}
\end{document}